\documentclass{article}

\usepackage[preprint]{neurips_2026}

\usepackage[utf8]{inputenc}
\usepackage[T1]{fontenc}
\usepackage{float}
\usepackage{hyperref}
\usepackage{url}
\usepackage{booktabs}
\usepackage{amsfonts}
\usepackage{microtype}
\usepackage{xspace}
\usepackage{multirow}

\hypersetup{
  pdfauthor={Cameron Manzo},
  pdftitle={Writing Style Similarity Reflects Academic Genealogy},
  pdfsubject={},
  pdfkeywords={}
}

\title{Writing Style Similarity Reflects Academic Genealogy}
\author{%
  Cameron Manzo \\
  RTX BBN Technologies \\
  \texttt{ctmanzo3@icloud.com}
}

\begin{document}
\maketitle

\begin{abstract}
As authorship attribution systems are increasingly deployed to detect
ghostwritten and AI-generated papers, their errors can support accusations
against legitimate authors. These systems conflate stylistic similarity with individual identity. Researchers, however, study under advisors, and inherit their stylistic
quirks. We build a corpus of arXiv authors with $\geq 2$ solo papers from the
Mathematics Genealogy Project graph, giving $5{,}803$ total authors and
$2{,}501$ ground-truth advisor-student pairings. Using embeddings from a fine-tuned model, advisors sit $39.9\%$ closer in cosine distance to their
students than a random same-field author does. Using two open models, we reproduce the effect at $12.6\%$ and $14.5\%$. \emph{Academic siblings}, two students of
one advisor who may never have met, sit $30.4\%$ closer across $8{,}360$
pairs, even when they studied at different institutions. Pairs who share only institution and field show negligible similarity. Given a closed-set attribution task over the same corpus, the system's errors occur on the true author's advisor, student, academic sibling, or lab mate $11$ times more often than chance.
\end{abstract}

\section{Introduction}
When an attribution system flags a student's paper as written by their advisor, the false positive is difficult to disprove during peer review. Whether such systems systematically produce these false
accusations depends on how much of a researcher's writing style is derived from others.

Advisors and students share subfields, departments, and native languages. Each of these factors inflates stylistic similarity independently of mentorship. Therefore, we
control for shared fields and institutions to discover a true effect.
Furthermore, these relationships should be verified; existing measurements of advisor influence infer such relationships
(Section~\ref{sec:related}). To address this limitation, we build our corpus
from the Mathematics Genealogy Project (MGP) graph, which has been curated for
ground truth.

Our contribution is a ground-truth based measurement of academic genealogy in writing
style. We link curated MGP records to authorship-attribution embeddings,
creating 2{,}501 advisor-student pairs and 14{,}398 total sibling pairs
(including 8{,}360 across different institutions). To our knowledge, we provide
the first stylistic analysis of academic siblings. We use these results to show
how this effect causes significant failures in closed-set attribution
tasks.

\section{Related Work}
\label{sec:related}

\paragraph{Writing style dynamics.}
\citet{lazebnik2025computational} modeled individual scholarly style evolution
across 13.7M publications, finding that trajectories stabilize around
publication 13 and that advisor influence is measurable in early-career work.
Advisors there are inferred as a scholar's most frequent co-author in their
first three publication years, and the analysis reaches no further than the
direct advisor link. \citet{danescu2011mark,danescu2012echoes} established that
speakers converge toward the style of high-status interlocutors in online
communities, providing a theoretical mechanism for the advisor effect we
measure. We extend both lines by measuring against curated genealogy records
and past the direct link to academic siblings.

\paragraph{Academic genealogy.}
Studies of the MGP~\citep{david2022academic} have traced how research topics
propagate through advisor lineages. We ask whether style follows a similar mechanism.

\paragraph{Authorship attribution.}
Authorship attribution systems map text to author embeddings and identify
authors by nearest-neighbor search in embedding space~\citep{pan2024}.
Purpose-built encoders such as LUAR~\citep{rivera2021luar} and
content-independent models such as
StyleDistance~\citep{patel2024styledistance} represent the state of the art.
These systems are built on the implicit assumption that each author's style is
independent---that knowing one author's style tells you nothing about
another's. We challenge that assumption using verified mentorship records.

\section{Data and Methods}
\label{sec:methods}

\paragraph{Corpus.}
We crawled 5{,}715 Mathematics Genealogy Project pages, yielding 28{,}662
distinct people, of whom 6{,}133 are eligible (those with $\geq$2
single-authored arXiv papers) and 5{,}803 have text and embeddings. We control
date to pre-2024 to avoid large changes in style from proliferation of LLMs.
Sole-author papers were used throughout so embeddings reflect a ground-truth
individual style. Author text is the \emph{abstracts} of each author's
sole-authored papers, taken from the arXiv metadata snapshot. This yields a mean 5.7 papers
per author. We strip affiliation strings before embedding, preventing them from inflating same-institution similarity. Because we use abstracts rather than full text, our measured effect is likely conservative. 
\paragraph{Embeddings.}
The main encoder (henceforth ``the tuned encoder'') is a Mistral-based model fine-tuned for authorship attribution following \citet{kandula2025alert}. Weights
and training data are not public. We split each paper into chunks, embed each chunk (1,536 dimensions), and average first to one vector per paper, then average those such that we get one vector per author. \emph{Similarity} is the cosine distance between two authors' vectors. 

\paragraph{Institution assignment.}
We consider institution to be each person's PhD-granting school as recorded on their MGP
page. This records
where an author \emph{received their PhD}, not where they currently work. This yields 3{,}609 of the 6{,}133 eligible people (59\%) across 459 unique institutions.

\paragraph{Advisor-student pairs.}
\label{par:mgpfix}
 Relationships are read directly from the MGP graph by numeric page ID. When resolving a person to their arXiv record, we discard names claimed by two different records. Pairs without embeddings for both members are dropped, leaving the advisor-student and sibling samples in Table~\ref{tab:labmate}. Sibling pairs are split by whether the two members trained at the same institution or different institutions. Siblings who trained at different institutions give us our cleanest test of style inherited purely from the advisor. A subset of 1,286 advisor pairs is analyzed by career phase in Appendix~\ref{sec:career} to see if this effect holds over time.

\paragraph{Statistical validation.}
We validate our result by replacing one member of each pair with a random author from
the same field and recomputing the mean distance. $p$-values are the fraction
of 1{,}000 draws at or below the observed value, so $p<.001$ is the smallest
possible value.

\paragraph{Open model replication.}
We replicate our findings with LUAR~\citep{rivera2021luar} (intended for
authorship representation) and StyleDistance~\citep{patel2024styledistance}
(content-independent). 

\section{Results}
\label{sec:results}

\subsection{Lab-Mate Similarity}
Once advisor-student and sibling links are removed, shared PhD institution on
its own does not produce a reliable stylistic effect. Under the author-level
permutation test, the pooled same-institution mean sits 3.2\% below its
permutation null under the tuned encoder across 57{,}250 pairs among 3{,}083
people. We find LUAR gives $-1.0\%$
($p=1.000$) and StyleDistance $-0.6\%$ ($p=0.997$). On the normalized scale
(Table~\ref{tab:normalized}) the tuned-encoder figure is 5.0\%
of the way from a random stranger toward the same author, against 61.7\% for
advisors.

\begin{table}[h]
\centering
\caption{Writing similarity by relationship, under the tuned encoder. Each row
is measured against its own permutation null (range 0.0807--0.0834), formed by
replacing one member of each pair with a random same-field author. \%cl = \%
closer than null. For the same-person value we use the distance between two
halves of the same author's papers; it is the 0 point of
Table~\ref{tab:normalized}}
\label{tab:labmate}
\begin{tabular}{lccc}
\toprule
Comparison & Mean cosine dist. & \%cl & $N$ pairs \\
\midrule
Same author & 0.0295 & --- & 3{,}136 \\
Advisor--student & \textbf{0.0501} & 39.9\% & 2{,}501 \\
Sibling, same institution & 0.0531 & 34.2\% & 6{,}038 \\
Sibling, different institution & 0.0578 & 30.4\% & 8{,}360 \\
Same PhD institution (\emph{lab mate}) & 0.0798 & 3.2\% & 57{,}250 \\
\bottomrule
\end{tabular}
\end{table}

\subsection{Advisor-Student Similarity}
\label{sec:advisor}
The permutation test gives $p<0.001$ for all three encoders. Both open models agree with the tuned encoder in direction and
significance at the pooled level, at smaller magnitude: 39.9\% under the tuned
encoder against 12.6\% under LUAR and 14.5\% under StyleDistance. Comparing embedding spaces is noisy, so we also report a
normalized scale (Table~\ref{tab:normalized}), on which the tuned encoder
and StyleDistance agree closely (61.7\% and 62.7\% of the way toward
same-person).

\begin{table}[h]
\centering
\small
\caption{Relationship distance vs.\ random same-field baseline. For advisor-student, we are field-agnostic, e.g., an advisor can publish in a different field than their student.
\%cl = \% closer than null.}
\label{tab:advisor}
\begin{tabular}{lrrr|rr|rr}
\toprule
& \multicolumn{3}{c|}{Tuned} & \multicolumn{2}{c|}{LUAR} &
\multicolumn{2}{c}{StyleDistance} \\
Relationship & \%cl & $p$ & $n$ & \%cl & $p$ & \%cl & $p$ \\
\midrule
Advisor--student         & \textbf{39.9\%} & $<$.001 & 2{,}501 & 12.6\% & $<$.001 & 14.5\% & $<$.001 \\
Sibling (all)           & \textbf{32.0\%} & $<$.001 & 14{,}398 & 7.3\% & $<$.001 & 6.3\% & $<$.001 \\
\quad same institution  & \textbf{34.2\%} & $<$.001 & 6{,}038 & 7.9\% & $<$.001 & 7.6\% & $<$.001 \\
\quad diff.\ institution & \textbf{30.4\%} & $<$.001 & 8{,}360 & 6.9\% & $<$.001 & 5.4\% & $<$.001 \\
\midrule
Same PhD institution    & 3.2\% & $<$.001 & 57{,}250 & $-$1.0\% & 1.000 & $-$0.6\% & .997 \\
\bottomrule
\end{tabular}
\end{table}

\subsection{Academic-Sibling Similarity}
\label{sec:sibling}
Two students of the same advisor write more similarly than chance under all
three encoders (Table~\ref{tab:advisor}). Siblings have no verified connection to each other, and
frequently no temporal overlap in the group at all. This pattern is best explained by a common mentor.

One may object that siblings are usually trained in the same department. However, splitting the pairs, we find siblings who trained at the \emph{same} institution sit
34.2\% below null and siblings who trained at \emph{different} institutions sit
30.4\% below null, a gap of under four points. Both open models show this trend (see Table~\ref{tab:normalized}). Siblings may also inherit their advisor's research program rather than
their style; the cross-field check (Appendix~\ref{sec:crossfield}) and
StyleDistance's content-independent training bound this confound but do not
eliminate it.

\subsection{Effect on an Attribution System}
\label{sec:attribution}
By themselves, the similarity results do not establish implications for authorship attribution systems. We
therefore run a closed-set attribution task and log the errors it makes.

For every author in our corpus with at least four papers, we take out one paper, build their
profile from the remainder, and rank all candidate authors by cosine distance
to the withdrawn paper. Pools are restricted to the same primary arXiv
category. With the tuned model, over 2{,}881 authors, the system ranks the true author first in 1{,}322/2{,}881 cases (45.9\%).

In 125 of these 1{,}559 cases (8.02\%) the
author the system picks is a social relative of the true author, which we consider an advisor,
a student, an academic sibling, or from the same PhD institution. Uniformly
distributed errors would produce this 0.73\% of the time, so this result occurs \textbf{11.0$\times$} more often than chance (
$p=6.1\times10^{-85}$). The direction is the same within fields. For those with usable size, the magnitudes are as follows:
cs 16.6$\times$, math 11.9$\times$, cond-mat 10.0$\times$, math-ph
7.8$\times$, hep-th 1.8$\times$.

\section{Discussion}
Our results indicate mentorship leaves a measurable trace in writing style, while shared institution alone does not. This similarity between advisors, students, and siblings can confuse authorship attribution systems at rates an order of magnitude greater than chance.

The decay of the advisor signal during an author's career (Appendix~\ref{sec:career}) is consistent with stylistic imprinting from advisors. Even if students choose advisors who write similarly to them initially, it cannot explain why this similarity decreases over time, consistent with an independent voice emerging.

Siblings need not have met or even publish in the same field, ruling out co-authorship or shared lexicon. Hence, a common advisor remains the clearest explanation for why they write so similarly.

For attribution systems, genealogical relationships are a confound worth
modeling explicitly: a deployed system should check whether its best
candidate is a known relative of its second-best before supporting an
accusation.

\paragraph{Limitations.}
\emph{Field diversity:} because MGP originates from math,  the corpus is 71\% math, 8\% cs, 4\% hep-th, 2\% stat,
with a remainder in 40 other arXiv categories.

\emph{Subfield clustering and
shared linguistic background:} topic confounds are bounded by StyleDistance's
content-independent training and the cross-field check (Appendix~\ref{sec:crossfield}). Appendix~\ref{sec:negcontrol}
shows linguistic background contributes 18.1\%, which we cannot parse from
mentorship.

\emph{Name matching:} resolving a person to their arXiv record is name-based;
discarding ambiguous names removes collisions inside the corpus but not with
outside arXiv authors who share a corpus member's name.

\begin{ack}
This research is based upon work supported in part by the Office of the
Director of National Intelligence (ODNI), Intelligence Advanced Research
Projects Activity (IARPA), via 2022-22072200003. The views and conclusions
contained herein are those of the authors and should not be interpreted as
necessarily representing the official policies, either expressed or implied,
of ODNI, IARPA, or the U.S. Government. The U.S. Government is authorized to
reproduce and distribute reprints for governmental purposes notwithstanding
any copyright annotation therein.
\end{ack}


\appendix

\section{Normalized Effect-Size Scale}
\label{sec:normalized}

\begin{table}[H]
\centering
\small
\caption{Effect size on a scale from 0 to 1, where 0 is the author's own cosine distance between two randomly split halves of the same author's papers and 1 by a random same-field author. LUAR is excluded as it gives no per-document vector.}

\label{tab:normalized}
\begin{tabular}{lrr}
\toprule
Relationship & Tuned Enc. & StyleDistance \\
\midrule
Advisor--student & 61.7\% & 62.7\% \\
Sibling, same institution & 53.9\% & 34.9\% \\
Sibling, different institution & 47.1\% & 23.5\% \\
Same PhD institution & 5.0\% & $-2.7\%$ \\
\bottomrule
\end{tabular}
\end{table}

\section{Negative Control: Name-Collision False Matches}
\label{sec:negcontrol}
We evaluate 2{,}712 unrelated same-field author pairs who share a surname using the same pipeline.
As a result, authors who share surnames sit 18.1\% closer in style than random pairs ($p<0.001$). Surnames are a relatively good proxy for linguistic background, which shapes English writing style. However, the effect with advisors from the same field (43.2\%, vs.\ 39.9\% when field-agnostic) and different-institution siblings (30.4\%) are much larger than the surname effect (18.1\%). Surname groups with more than 40 corpus members are excluded from this
control, so 18.1\% is likely a lower bound. Regardless, we report this phenomenon as a limitation.

\section{Cross-Field Robustness Check}
\label{sec:crossfield}
The 473 cross-field advisor pairings, who write in different fields, sit
26.7\% below their null ($p<0.001$), compared to 43.2\% for same-field
pairs. If the phenomenon were purely topic-driven, this effect would not occur.

\section{Career-Phase Analysis}
\label{sec:career}
Using the 1{,}286 pairs whose student has at least four dated papers, we sort
each student's papers by arXiv date and pool the earliest half into an
early-career profile and the latest half into a late-career profile, then for each find the cosine distance
against the advisor's profile.

We find early-career mean distance is
0.0526 and late-career mean is 0.0563, a change of $+0.0037$ ($+7.1\%$; 95\%
bootstrap CI $[+0.0028, +0.0047]$). This refutes the idea that students choose advisors who already write similarly to themselves. Advisors seem to implicitly impose their style upon their mentees early in their careers, which is consistent with when an advisor would be active.

\end{document}